# Quasi Error-free Text Classification and Authorship Recognition in a large Corpus of English Literature based on a Novel Feature Set

Arthur M. Jacobs[1,2] & Annette Kinder[3]

Author Note

1 Experimental and Neurocognitive Psychology, Freie Universität Berlin, Germany

2 Center for Cognitive Neuroscience (CCNB), Freie Universität Berlin, Germany

3 Learning Psychology, Freie Universität Berlin, Germany

Correspondence: Arthur M. Jacobs

Department of Experimental and Neurocognitive Psychology, Freie Universität Berlin,

Habelschwerdter Allee 45 , D-14195 Berlin, Germany.

Email: ajacobs@zedat.fu-berlin.de

**Abstract**

The *Gutenberg Literary English Corpus (GLEC)* provides a rich source of textual data for research in digital humanities, computational linguistics or neurocognitive poetics. However, so far only a small subcorpus, the *Gutenberg English Poetry Corpus,* has been submitted to quantitative text analyses providing predictions for scientific studies of literature. Here we show that in the entire GLEC quasi error-free text classification and authorship recognition is possible with a method using the same set of five style and five content features, computed via style and sentiment analysis, in both tasks. Our results identify two standard and two novel features (i.e., type-token ratio, frequency, sonority score, surprise) as most diagnostic in these tasks. By providing a simple tool applicable to both short poems and long novels generating quantitative predictions about features that co-determine the cognitive and affective processing of specific text categories or authors, our data pave the way for many future computational and empirical studies of literature or experiments in reading psychology.

Text corpora are an indispensable tool for scientific studies in many fields such as natural language processing (NLP), digital humanities or neurocognitive poetics[1-5] offering the textual basis for tackling numerous novel research questions, such as why *Molière* most likely did write his plays[6], whether *Cynewulf* and *Andreas* are stylistically associated[3], whether different dramatic subgenres in French dramas of the classical age have distinctive dominant topics[7], what the dominant shapes of the emotional arcs of stories are[8], how happy people were during previous centuries[9], or which topics dominate the 154 Shakespeare sonnets[10] and which text features mainly control readers eye movements during reading such sonnets[11,14]. Especially empirical studies investigating not only texts themselves –as in NLP or stylometrics– but behavioral or neuronal reader responses to texts like in the sonnet studies[11,14] or neuroimaging reading experiments on texts from the *Harry Potter* series[12] can benefit from results on quantitatively analysed text corpora because they provide predictions based on the number and type of salient features systematically discriminating one type of text from another, or inducing one dominant mood in a given poem, for example[13]. The *GLEC* is such a tool offering plenty of novel possibilities for stylometric and empirical studies of literature with a high potential for providing benchmarks, for example on authorship recognition performance[1]. It contains ~900 novels, 500 short stories, 300 tales and stories for children, 200 poetry collections, poems and ballads, 100 plays, as well as 500 pieces of non-fiction, e.g. articles, essays, lectures, letters, speeches or (auto-)biographies, with *~12 million* sentences and *250 million* words from a wide range of authors such as Austen, Byron, Coleridge, Darwin, Dickens, Einstein, Eliot, Poe, Twain, Woolf, Wilde, or Yeats (see examples in Table 1). Previous work presented the results of stylometric, topic and sentiment analyses only for a subset of *GLEC*, the ~120 texts from the *Gutenberg English Poetry Corpus* motivating several behavioral experiments investigating eye movement control during the reading of Shakespeare sonnets[11,14]. The present analyses extend this work to the whole *GLEC* providing answers to two basic questions that can guide future scientific studies of literature and reading psychology. First, which text features are most diagnostic in determining broad text categories (e.g., essay, novel, or poetry), and, second, which text features optimise authorship recognition.

The number of text features that in principle co-determine literature reception or allow a reader to discern an essay and a fictive story is unknown and depends on the text-context-reader trias, i.e. a complex nonlinear dynamic system in which reader-specific personality variables can interact with a myriad of interdependent hierarchical text and context features whose effects can only be predicted by powerful machine learning models fed with data from advanced quantitative narrative and sentiment analysis tools[5,15]. Since Jakobson and Lévi-Strauss'[16] seminal poetics study numerous studies in stylometrics, computational linguistics or reading psychology have dealt with diverse subsets of sublexical, lexical or supralexical features more or less successful in authorship recognition, sentiment analysis performance, or the prediction of human reading behavior[1,2,4,5,9,15]. A central issue in this research concerns the optimal combination of style and content features for performing different tasks like authorship recognition or eye movement prediction. Prominent style or surface features used both in text or genre classification and behavioral experiments are number and frequencies of tokens, sentence length, or type-token ratio/TTR[18]. More recent work also pointed to the usefulness of affective semantic (content) features for text genre classification[4,19] or the prediction of human text liking ratings[20]. Typical affective semantic (content) features are lexical valence, arousal, or discrete emotions like happiness or disgust[4,21,22]. Due to recent advances in computational linguistics and machine learning using large numbers of text features and their interactions is no longer a problem. However, the question which combination of style and content features is optimally diagnostic for text classification, authorship recognition or human behavior prediction has hardly been tackled and thus remains open. Here we demonstrate –as far as we know, for the first time– that the same combination of a few easy-to-compute style and content features achieves quasi-perfect performance in both text classification and authorship recognition.

**Table 1 here.** 60 example texts from *GLEC,* 10 from each category.

| CHILDREN AND YOUTH LITERATURE | ESSAYS |
|---|---|
| Andrew Lang.Tales of Troy and Greece | Bertrand Russell.The Analysis of Mind |
| Baronness Orczy.The Scarlet Pimpernel | Charles Darwin.On the Origin of Species 6th Edition |
| Beatrix Potter.The Tale Of Peter Rabbit | George Eliot.The Essays of George Eliot1 |
| Edward Stratemeyer.The Rover Boys in the Land of Luck | John Locke.An Essay Concerning Humane Understanding |
| Jacob Abbott.Cleopatra | John Stuart Mill.A System Of Logic |
| James Matthew Barrie.Peter Pan | Lewis Carroll.Symbolic Logic1 |
| Louisa May Alcott.Rose in Bloom | Michael Faraday.Experimental Researches in Electricity |
| Lyman Frank Baum.The Wonderful Wizard of Oz | Sir Isaac Newton.Opticks |
| R M Ballantyne.Away in the Wilderness | Sir Winston Churchill.Liberalism and the Social Problem |
| Thornton Waldo Burgess.Mrs. Peter Rabbit | William Butler Yeats.Discoveries |
| **NOVELS** | **PLAYS** |
| Bram Stoker.Dracula | George Bernard Shaw.Pygmalion |
| Charles Dickens.Oliver Twist | John Dryden.All for Love |
| D H Lawrence.Women in Love | John Galsworthy.A Bit O Love |
| Daniel Defoe.The Life and Adventures of Robinson Crusoe | Oscar Wilde.Lady Windermeres Fan |
| Edgar Rice Burroughs.Tarzan of the Apes | Richard Brinsley Sheridan.The Rivals |
| George Eliot.Middlemarch | William Butler Yeats.The Hour Glass |
| Herman Melville.Moby Dick | William Dean Howells.The Sleeping Car |
| Jane Austen.Emma | William Shakespeare. Hamlet-Prince of Denmark |
| Sir Arthur Conan Doyle.The Hound of the Baskervilles | William Shakespeare. Romeo And Juliet |
| Winston Churchill.Coniston | William Shakespeare.The Tempest |
| **POETRY** | **SHORT STORIES** |
| Alexander Pope.The Poetical Works | Charles Dickens.Holiday Romance |
| D H Lawrence.Amores | George Eliot.Brother Jacob |
| Elizabeth Barrett Browning.The Poetical Works | Henry James.The Chaperon |
| John Keats.Endymion1 | Jane Austen.Love And Friendship |
| John Milton.The Poetical Works of John Milton1 | Joseph Conrad.Youth |
| Lord Byron.Byrons Poetical Works Vol. 1 | Lucy Maud Montgomery.Short Stories 1896 to 1901 |
| P B Shelley.The Complete Poetical Works | Mark Twain.The Facts Concerning The Recent Carnival Of Crime In Connecticut |
| Samuel Taylor Coleridge.The Complete Poetical Works1 | O Henry.Rolling Stones |
| William Butler Yeats.The Wild Swans at Coole | Oscar Wilde.Shorter Prose Pieces |
| William Shakespeare.sonnet32 | Rudyard Kipling.Indian Tales |

Text classification is a standard task in machine learning assisted NLP usually tackling the problem of correctly labeling test texts from newspapers or internet sources into genre categories like business or politics, or literary texts into categories like adventure or romance. An early influential NLP study[18] of the ~1 million word *Brown corpus*[2] used easy-to-compute style features requiring no tagging (e.g., sentence length, TTR) fed into a simple feedforward neural net (a Multi-Layer Perceptron/MLP) and achieved maximum accuracies of about 80% for *binary* classifications such as whether a text is a narrative or not. More recent work on literary texts sampled from the Gutenberg project introduced affective semantic content features like joy or anger for genre classification (e.g., western or romance) fed into a decision tree algorithm which achieved a maximum *multiclass* classification accuracy of ~80% for the category science fiction[4]. Applying an ensemble of seven machine learning classifiers (e.g., MLP, convolutional neural net) that combined their predictions based on either simple bag-of-word features (i.e., the 5000 most frequent words) or on emotion labels to ~2000 texts from the Gutenberg project another recent study yielded a maximum performance of 94% for five categories (adventure,humor,mystery,romance, science fiction)[19]. The affective semantic features in both studies were computed using a wordlist based sentiment analysis which requires matching the words of a text with an annotated dictionary providing human emotion ratings. Wordlist based tools have several limitations, though, that can be overcome by using vector space model based sentiment analysis tools like *SentiArt*[15,20].

Here we chose ~2700 texts from six broad *GLEC* categories (children and youth literature/CYL, essays/ESS, novels/NOV, plays/PLA, poems/POE and short stories/STO) as a testing ground for an innovative text classification and authorship recognition study which used a novel set of five easily computable style and five content features never before applied for these purposes. The five style features are: i) FREQ, i.e. the mean token log frequency, a *lexical* measure indicating text familiarity and influencing readability[11,14]; three supralexical measures: ii) TTR, i.e. the type/token ratio (number of unique words / number of tokens), a measure that is often used as an indicator of linguistic complexity, poetic quality, or aesthetic success[5], iii) SSI, the sentence syllable index, i.e. the product of number of tokens (i.e. sentence length) by mean number of syllables per sentence, a measure inspired by traditional readability indices[25], larger values indicating lower readability; and iv) NSENTS, i.e. the number of sentences per text; and, finally, one sublexical measure: v) SONSCORE, i.e. the mean sonority score per text, a simplified index based on a words phonemes sonority hierarchy that has proven to be diagnostic in predicting human word beauty ratings[23] or the literariness of metaphors[24] (see Methods for further details). The five content features were: AAP, DISGUST, FEAR, HAPPINESS, and SURPRISE. AAP refers to the mean affective-aesthetic potential of a text, a novel measure based on the semantic relatedness between the text tokens and each of 120 words of a label list –computed via the word2vec algorithm[26] (see Methods). This measure successfully predicted human word beauty ratings[23], human valence ratings, and liking ratings for text passages[20], but was not yet applied to the tasks of text classification and authorship recognition. The remaining four affective semantic features (DISGUST, FEAR, HAPPINESS, SURPRISE) refer to a words semantic relatedness – computed via word2vec– with either of these four discrete emotion concepts (see Methods). For example, a text with a high SURPRISE value would feature many words semantically related to surprise.

**Results**

Table 2a-c summarizes the results of Experiment 1 on text classification and Table 3a-b those of Experiment 2 on authorship recognition. In both experiments the same 10 features were computed for all 2722 texts of *GLEC*, and then fed into a standard MLP classifier that was trained on a part of these texts in a 5-fold cross-validation procedure to predict both text category and authorship for the remaining texts in the test sets (see Methods). As a control we also computed the (mis-)classification score for an alternative standard classifier (Naïve Bayes/NB) which implements the assumption that the 10 style and content features do not interact. The topmost Table 2a shows a near perfect overall multiclass classification accuracy of the MLP (misclassification rate < .013) for the test set of 544/2722 texts (NB misclassification rate for test set = .33; see Methods for explanation of fit indices). The normalized confusion matrix in Table 2b also shows perfect or near perfect rates for all but one category: only STO were confused in about 3% of the cases with NOV which share plenty of style and content features with them. In two cases, a story was also confounded with an essay. Novels were mistakenly predicted by the MLP to be stories in 2/544 cases. Table 2c shows the feature (variable) importances (FI) for all six text categories sorted by their overall importance (rightmost column). They present an interesting mix for the overall top four (tie between ranks 3 and 4): a traditional supralexical style feature (TTR) is slightly more diagnostic than a novel sublexical content feature (SONSCORE), followed by a novel lexical content feature (SURPRISE) and a novel supralexical style feature (SSI). The remaining seven features were all important, too (FIs >.1[27]), with rather small differences between them.

For individual text categories both the ranking and FI values can deviate from the overall pattern. Thus, for ESS the top ranking feature was SURPRISE (FI = .33), while SSI dominated for PLA and SONSCORE for POE. Not surprisingly, NSENTS top ranked for NOV and second for POE –the longest and shortest texts in *GLEC*, respectively.Thus, altogether affective semantic and style features are about equally important in co-determining the category of a *GLEC* text. Notably, SONSCORE was among the top three FI values in 4/6 individual text categories (POE, ESS, STO, CYL). This is strong evidence that this sublexical phonological feature which has been shown to affect metaphor and poetry processing in human readers[11,14,20] also should play a role in future NLP and empirical studies of literature –as should the remaining features, none of which exhibited an unimportant overall effect (<.1) in these data.

**Table 2a-c.** Results for text classfication.

**a) Model Fit Indices Text Classification**

| Measures | MLP | NB |
|---|---|---|
| Generalized $R^2$ | 0,992 | - |
| Entropy $R^2$ | 0,945 | - |
| RMSE | 0,143 | - |
| Mean Abs Dev | 0,075 | - |
| Misclassification Rate | 0,012 | .33 |
| -LogLikelihood | 49,31 | - |
| Sum Freq | 544 | - |

**b) Confusion Matrix (%) Text Classification**

| text category actual / predicted | CYL | ESS | NOV | PLA | POE | STO |
|---|---|---|---|---|---|---|
| CYL | 1,000 | 0,000 | 0,000 | 0,000 | 0,000 | 0,000 |
| ESS | 0,000 | 1,000 | 0,000 | 0,000 | 0,000 | 0,000 |
| NOV | 0,000 | 0,006 | 0,99 | 0,000 | 0,000 | 0,037 |
| PLA | 0,000 | 0,000 | 0,000 | 1,000 | 0,000 | 0,000 |
| POE | 0,000 | 0,000 | 0,000 | 0,000 | 1,000 | 0,000 |
| STO | 0,000 | 0,009 | 0,027 | 0,009 | 0,000 | 0,955 |

**c) Feature Importance (0 <= FI <= 1) Text Classification by Text Category**

| Feature | CYL | ESS | NOV | PLA | POE | STO | Importance Overall |
|---|---|---|---|---|---|---|---|
| TTR | 0,441 | 0,207 | 0,16 | 0,194 | 0,203 | 0,365 | 0,262 |
| SONSCORE | 0,153 | 0,23 | 0,129 | 0,23 | 0,309 | 0,254 | 0,218 |
| SURPRISE | 0,15 | 0,328 | 0,17 | 0,218 | 0,069 | 0,298 | 0,205 |
| SSI | 0,124 | 0,121 | 0,106 | 0,582 | 0,1 | 0,197 | 0,205 |
| NSENTS | 0,344 | 0,061 | 0,211 | 0,149 | 0,217 | 0,137 | 0,187 |
| FREQ | 0,111 | 0,13 | 0,13 | 0,304 | 0,067 | 0,217 | 0,16 |
| HAPPINESS | 0,113 | 0,089 | 0,066 | 0,326 | 0,114 | 0,13 | 0,14 |
| DISGUST | 0,109 | 0,092 | 0,1 | 0,224 | 0,049 | 0,166 | 0,123 |
| FEAR | 0,132 | 0,094 | 0,114 | 0,154 | 0,097 | 0,121 | 0,119 |
| AAP | 0,114 | 0,113 | 0,084 | 0,224 | 0,055 | 0,141 | 0,122 |

Color code: Yellow = top rank, Green = 2nd, Cyan = 3rd.

The second experiment using *GLEC* concerned authorship recognition (attribution), a notably more difficult task than the previous sixfold text classification, given that >100 different authors appear at least five times in *GLEC*, many of which contributing to several text categories and offering a rich creative variety of style and content features. Since Mosteller and Wallaces classical study[28], authorship recognition is a pivotal task in NLP. Despite considerable progress in stylometrics and ML techniques it remains a challenge in computational research on literature focusing on issues such as the choice of optimal features and classifiers[17] or minimal sample length for obtaining stable results[29]. Here we tested i) how well the MLP performed in recognizing the altogether 113 different authors of the present *GLEC* subset, ii) whether success rate depended on text category, and iii) which of the 10 features were the most diagnostic. Table 3a and b show author recognition scores and FI values for each of the six text categories, as well as for all text categories aggregated (rightmost column). Note that the latter condition represents the most difficult test in that the MLP must perform a conjunctive learning task since many authors appear in different text categories. As is obvious from Table 3a, within-text category authorship recognition was perfect for the MLP classifier in all but two categories (misclassification rates: NOV = .006; STO = .009), while the NB misclassification rates ranged from a low of .14 for CYL to a high of .55 for ESS. Among the authors who were confounded with each other once by the MLP were, for example, Bram Stoker and Daniel Defoe, Edgar Rice Burroughs and Zane Grey, George Eliot and Charlotte Mary Yonge, PG Wodehouse and Bret Harte, or Walt Whitman and John Ruskin. The still excellent across-category (ALL 6) performance with <4% misclassifications for >100 authors with >2500 texts underlines the power of the present mixed feature set. To our knowledge, this is the first demonstration of such a combined, quasi-perfect literature classification. Table 3b shows the different FI rankings for the seven conditions sorted according to the FI score for the All 6 across-category case. Among the top four features the same two novel content features as for text classification were found (SURPRISE and SONSCORE), while mean word frequency, a standard style feature strongly affecting reading time[11,14], was the most diagnostic when authors had to be recognized across different text categories. Looking at the individual text categories, the style feature SSI was most important for authorship recognition, while the content feature AAP ranked three times among the top important ones for ESS, NOV and PLA. This underlines the status of this latter variable that has already proven to be an important predictor of human liking ratings for varying verbal materials (words, passages)[20,23] as a promising multi-purpose feature in various NLP tasks or behavioral experiments. It should be noted, though, that the FI values in the ALL 6 condition of Table 3b are all pretty close together indicating that no single feature is dominantly diagnostic in this highly difficult task, but that it is their dynamic interplay which warrants the excellent performance of the MLP. Thus, just like in text classification, affective semantic and style features were about equally important in co-determining the authorship of a *GLEC* text.

**Table 3a-b.** Results for authorship recognition.

**a) Model Fit Indices Authorship Recognition**

| Measures | CYL | ESS | NOV | PLA | POE | STO | ALL 6 |
|---|---|---|---|---|---|---|---|
| Generalized $R^2$ | 0,999 | 0,999 | 0,999 | 1 | 0,999 | 0,999 | **0,999** |
| Entropy $R^2$ | 0,999 | 0,973 | 0,98 | 0,999 | 0,98 | 0,93 | **0,955** |
| RMSE | 0,000 | 0,13 | 0,122 | 0,000 | 0,088 | 0,262 | **0,233** |
| Mean Abs Dev | 0,000 | 0,9 | 0,067 | 0,000 | 0,05 | 0,199 | **0,14** |
| Misclassification Rate | 0 | 0 | 0,006 | 0 | 0 | 0,009 | **0,036** |
| -LogLikelihood | 0,000 | 10,8 | 12,84 | 0,000 | 3,59 | 27,416 | **105,49** |
| Sum Freq | 73 | 109 | 162 | 25 | 65 | 110 | **544** |

Misclassifications rates for NB. CYL: .18, ESS: .47, NOV: .37, PLA: .23, POE: .29, STO: .52, ALL 6: .51.

**b) Feature Importance (0 <= FI <= 1) Authorship Recognition**

| Feature | CYL | ESS | NOV | PLA | POE | STO | ALL 6 |
|---|---|---|---|---|---|---|---|
| FREQ | 0,135 | 0,42 | 0,267 | 0,241 | 0,411 | 0,341 | **0,407** |
| SURPRISE | 0,149 | 0,372 | 0,256 | 0,268 | 0,268 | 0,34 | **0,403** |
| SONSCORE | 0,159 | 0,4 | 0,222 | 0,161 | 0,357 | 0,359 | **0,403** |
| SSI | 0,237 | 0,506 | 0,428 | 0,36 | 0,598 | 0,415 | **0,397** |
| TTR | 0,199 | 0,41 | 0,322 | 0,245 | 0,307 | 0,272 | **0,393** |
| AAP | 0,175 | 0,505 | 0,313 | 0,298 | 0,28 | 0,286 | **0,393** |
| HAPPINESS | 0,186 | 0,365 | 0,216 | 0,298 | 0,258 | 0,307 | **0,36** |
| FEAR | 0,166 | 0,299 | 0,182 | 0,156 | 0,162 | 0,229 | **0,33** |
| DISGUST | 0,148 | 0,355 | 0,282 | 0,17 | 0,25 | 0,305 | **0,32** |
| NSENTS | 0,218 | 0,288 | 0,227 | 0,307 | 0,29 | 0,214 | **0,26** |

Color code: Yellow = top rank, Green = 2nd, Cyan = 3rd.

To cross-check the validity and generality of the present approach and test whether the same 10 features are useful not only for classifying broad text categories but also more refined literary genres like 'adventure' or 'romance', we sampled ~2000 texts from the Gutenberg genre corpus[19]. We analyzed these texts following exactly the same procedure as for the two experiments on *GLEC*: 649 'adventure' texts from 33 authors, 476 'mystery' texts (20 authors), 472 'romance' (11 authors), and 377 'western' stories (20 authors), using only texts from authors who appeared at least five times in a given genre. The results showed perfect authorship recognition scores for all four genres (all misclassification rates = 0) and an excellent multiclass genre misclassification rate of .038 (NB = .55). The top5 FI values in genre classification were: AAP = .297, SONSCORE = .283, SSI = .254, TTR = .249, and NSENTS = .236. Thus, the top3 were all novel features lead by an affective semantic one. This third study thus strengthens the evidence in favor of the versatility and general validity of the present feature set for both genre and authorship classification.

**Discussion**

In sum, our results show that the same set of only 10 features which can easily be computed –or even simpler, looked-up in a table (see Methods)– for any literary text, be it a short Shakespeare sonnet or a long Dickens novel, is highly effective for both determining its broad category (e.g., essay, short story, novel) and its author. This is the first time, to our best knowledge, that the same combination of traditional (and novel) style, and novel affective semantic features, is successfully applied to two of the most popular NLP tasks, text classification and authorship recognition. Our FI data in both tasks strongly suggest that style and content features both play a role in both tasks, their exact ranking varying as a function of text category. Thus a style feature may dominate in correctly determining that a text belongs to the poem category, while a content feature is more diagnostic in telling that a text is an essay. Similarly, deciding whether a novel was written by Dickens may depend on another most important feature than telling that an essay stems from Darwin. The results from the 3$^{rd}$ study on genre classification strengthen the evidence by showing that our 10 features also successfully can be applied to finer genre categories.

We believe that this flexible highly interactive and hierarchical processing of a limited set of text features also underlies human judgements of text category or authorship. Thus, the present results are of considerable interest to researchers in psychology or neurocognitive poetics[30,31] who may use our corpus and tools for designing experiments on human literature reception, testing, e.g., the usefulness of the present 10 feature set for recognizing the style of an author, or for predicting the likeability of a narrative –which depends on the AAP feature[20]. Our evidence for the efficiency of this limited feature set does not prove that it is generally optimal and will work similarly well for other, more modern texts (in other languages) or that human readers really explicitly or implicitly rely on some or all of these features; like all computer models, the present ones can only provide sufficiency, but not necessity analyses.

However, this data considerably challenges future studies in NLP, digital humanities or computational stylistics by providing a serious benchmark against which to test alternative feature sets.

The present approach can also serve as a model applicable to other corpora in different languages or from different time periods, such as the German *childLex* corpus[32]. It could be used, for instance, for testing the assumption that children and adolescents of different age groups differentially develop sensitivities for text features like TTR or AAP, likely depending on their overall reading experience, literary background and personal preferences. Our feature set (e.g., HAPPINESS, DISGUST, AAP or SONSCORE) can further be used in predicting the success of texts as measured by Goodreads or similar book rating agencies. A recent literature review on the possible causes of liking verbal materials ranging from single words to entire poems and books[33] suggests that a main driving force behind human liking ratings is the extent to which such high-dimensional stimuli are associated with the basic emotions joy/happiness and disgust, two important features in the present set.

**Methods**

**Corpus and text pre-processing.** Our set of 2722 texts was taken from the ~3000 texts of GLEC[1] which are available from the authors in .txt format. Only texts from writers that occurred at least five times in one of the six text categories were selected. This was necessary to enable the standard 5-fold cross-validation procedure for the machine learning part. All texts were pre-processed using standard NLP routines available from the open access NLTK library[34]. In particular, every sentence in each text was POS-tagged and lemmatized using the treetagger routine[35] (https://www.cis.uni-muenchen.de/~schmid/tools/TreeTagger/) excluding stopwords form the NLTK list and only content words (nouns, verbs, adjectives and adverbs) were retained. Using the *SentiArt*[15,20] tool each word in each sentence of each text was quantified in terms of a number of features such as word length, log frequency of occurrence in *GLEC*, AAP etc. and feature values were then aggregated across sentences and texts. For example, the AAP feature for a given text is the mean of all word AAP values of that text. All feature values for a given word in a given sentence can directly by retrieved from the *GLEC SentiArt* .xlxs table available from the authors (and eventually on github). This table provides basic feature values for ~250k words from GLEC covering 100% of the words appearing in the present texts. Thus, users simply can do a cross-table match for obtaining the 10 feature values for the words of a test text. It is likely, though, that our table does not cover 100% of the words from more modern English texts, but this problem could be solved by merging these texts (if publically available) with *GLEC*.

**Computation of 10 features**. The word features provided in the *GLEC SentiArt* table were computed as follows.

1. FREQ. The logarithm of the frequency of occurrence of a word in the ~250 million of *GLEC*.

2. NSENTS. The number of sentences per text, as computed by NLTKs sentence tokenizer.

3. TTR. The ratio of the number of tokens and the number of types (i.e. unique tokens), as computed by NLTKs word tokenizer. To compensate for text length log(TTR) was used[5].

4. SSI. The sentence syllable index, i.e. the product of number of tokens and mean number of syllables per sentence, the latter being computed via the novel *syllapy* library (https://github.com/mholtzscher/syllapy). Larger SSI values indicate that a sentence is theoretically less readable resulting in longer reading times.

5. SONSCORE. The sonority score of a word is the sum of its phonemes sonority hierarchy values divided by the square root of its length in letters. To compute this without requiring a text-to-speech conversion, a simplified grapheme heuristic is used based on the following sonority hierarchy of English phonemes, yielding 10 ranks: [a] > [e o] > [i u j w] > [ɾ] > [l] > [m n ŋ] > [z v] > [f θ s] > [b d g] > [p t k][23]. For example, the word 'art' gets a SONSCORE of 1 x 10 [a] + 7 × 1 [r] + 1 × 1 [t] = 18/SQRT (3) = 10.39, while 'game' (with a silent e) obtains 8.5. Higher SONSCORE values increase a words fixation probability and total fixation duration[11] and correlate with word beauty[23] and metaphor literariness[24].

6. HAPPINESS. The mean of the semantic relatedness (or similarity) values between each word in the text and the label word 'happiness'. This is computed using the *GLEC* vector space model/VSM which was created by the authors applying the *fasttext* algorithm (https://fasttext.cc/) to the ensemble of texts in *GLEC*. The final model contained a 500d *skipgram* vector for each word from *GLEC* ordered by frequency of occurrence. The cosine of the two word vectors gives the semantic relatedness value[23]. The GLEC VSM is available from the authors as a .txt file (and eventually on github). Within the context of NLP sentiment analyses, a text's overall HAPPINESS score is assumed to indicate that the text has a higher theoretical probability to induce positive emotional responses in readers. A similar index has recently been shown to predict reliable patterns in historical subjective wellbeing[9].

7.-9. FEAR, DISGUST, SURPRISE. The procedure is the same as for HAPPINESS, replacing the label word correspondingly. Higher FEAR or DISGUST values suggest that texts more likely induce negative feelings, while higher SURPRISE values indicate that readers have a higher probability of positive feelings associated with surprise.

10. AAP. The average semantic relatedness between each word in the text and *m* positive labels (*lpos*_1-60, = affection, amuse, ..., unity) minus the average similarity between each word and *n* negative labels (*lneg*_1-60, = abominable, ..., ugly). The labels were those published in earlier papers[5,23]. The semantic relatedness between a test word and each of the 120 labels is computed as for HAPPINESS. Mean values are then computed for the 60 positive and 60 negative labels and the difference between the means gives the final AAP value. Higher AAP values indicate a texts higher potential for evoking positive affective responses including aesthetic feelings of liking and beauty[5,23].

This reduced set of 10 features was obtained from a larger set of text features automatically computed by *SentiArt* for each *GLEC* text by selecting the five most important style and content features as indicated by the *predictor screening* tool of the JMP 14 Pro statistical software (https://www.jmp.com/de_de/software.html).

**Machine learning**. The classifications tasks were all performed using the *Predictive Modeling Platform* of JMP 14 Pro successfully employed in previous papers[10,11,14,33]. Two standard classifiers from the platform were used, NB and Neural Networks/MLP. The MLP had one hidden layer with 100 units using the nonlinear hyperbolic tangent (TanH*)* activation function, and a 2nd hidden layer with 25 units using the linear identity (Lin) activation function. It used *boosting*, i.e. the process of building a large additive neural network model by fitting a sequence of smaller models. Each of the smaller models is fit on the scaled residuals of the previous model. The models are combined to form the larger final model. The process uses validation to assess how many component models to fit, not exceeding the specified number of models. The *boosting* parameter for number of models was set to 10. The (initial) learning rate parameter was set to 0.1. The tours parameter was set to 10 iterations for each task, the penalty parameter was set to square. The cross-validation procedure was k-fold with k = 5. The random seed parameter was always set to 1 ensuring that the present results can be exactly reproduced.

The *model fit and feature importance (FI) measures* are explained in the following.

| Measure | Definition |
| --- | --- |
| Entropy $R^2$ | 1-Loglike(model)/Loglike(0) |
| Generalized $R^2$ | $(1-(L(0)/L(model))^{(2/n)})/(1-L(0)^{(2/n)})$ |
| RMSE | $\sqrt{\Sigma(y[j]-\rho[j])^2/n}$ |
| Mean Abs Dev | $\Sigma |y[j]-\rho[j]|/n$ |
| Misclassification Rate | $\Sigma (\rho[j]\neq\rho Max)/n$ |
| -LogLikelihood | $\Sigma -Log(\rho[j])/n$ |
| Sum Freq | n |

**Entropy $R^2$.** One minus the ratio of the negative log-likelihoods from the fitted model and the constant probability model. It ranges from 0 to 1.

**Generalized $R^2$.** A measure based on the likelihood function *L*, scaled to have a maximum value of 1. The value is 1 for a perfect model, and 0 for a model no better than a constant model. The measure simplifies to the traditional $R^2$ for continuous normal responses in the standard least squares setting. Generalized $R^2$ is also known as the Nagelkerke or Craig and Uhler $R^2$, which is a normalized version of Cox and Snells pseudo $R^2$.

**RMSE.** The root mean square error, adjusted for degrees of freedom. The differences are between 1 and p, the fitted probability for the response level that actually occurred.

**Mean Abs Dev.** The average of the absolute values of the differences between the response and the predicted response. The differences are between 1 and *p*, the fitted probability for the response level that actually occurred.

**Misclassification Rate.** The rate for which the response category with the highest fitted probability is not the observed category.

**-LogLikelihood.** The negative of the log-likelihood.

**Sum Freq.** The number of observations.

In the current study, **FIs** were computed as the total effect of each predictor (feature, variable) as assessed by the *dependent resampled inputs* option of JMP14 Pro. The total effect is an index quantified by sensitivity analysis, reflecting the relative contribution of a feature both alone and together with other features[36]. This measure is interpreted as an ordinal value on a scale of 0 to 1, **FI** values > 0.1 being considered as important[27].

**Reporting Summary.** Further information on research design is available in the Nature Research Reporting Summary linked to this article.

**Data availability.** For reproducing the statistical/machine learning analyses of the main data table, the JMP14 Pro software (not free) is necessary. A .xlxs version of the main table will be made freely and publicly available, though, at github or alternative outlets for users who wish to analyze the data with alternative software. A table listing the 2722 texts and the texts themselves will also be made freely and publicly available and can be obtained by the authors on demand, as can be the original log files from JPM14 Pro for the machine learning analyses (in .pdf format).

**Code availability.** All custom code will be made freely and publicly available at github or alternative outlets (e.g., a special server of FU Berlin).

# References

1 Jacobs AM (2018) The Gutenberg English Poetry Corpus: Exemplary Quantitative Narrative Analyses. *Front. Digit. Humanit*. 5:5. doi: 10.3389/fdigh.2018.00005

2 Kucera, H., & Francis, W. N. *Computational analysis of present-day American English.* Providence: Brown University Press, 1967.

3 Neidorf, L., Krieger, M. S., Yakubek, M., Chaudhuri, P., & Dexter, J. P. (2019). Large-scale quantitative profiling of the Old English verse tradition. *Nat Hum Behav*, 3(6), 560–567. Retrieved from http://www.nature.com/ articles/s41562-019-0570-1

4 Samothrakis, S., & Fasli, M. (2015). Emotional sentence annotation helps predict fiction genre. *PloS one* 10(11):e0141922.

5 Jacobs, A. M. (2018). (Neuro-)Cognitive Poetics and Computational Stylistics. *Scientific Study of Literature*, 8:1, pp. 165-208. https://doi.org/10.1075/ssol.18002.jac.

6 Cafiero, F. & Camps, J. P. (2019). Why Molière most likely did write his plays, *Science Advances*, 27 Novembre 2019, DOI : 10.1126/sciadv.aax5489

7 Schöch, C. (2017). Topic modeling genre: an exploration of french classical and enlightenment drama. *Digit. Human. Q.* doi: 10.5281/zenodo.166356.

8 Reagan, AJ, Mitchell, L, Kiley, D, Danforth, CD, & Dodds, PS (2016). The emotional arcs of stories are dominated by six basic shapes. *EPJ Data Science* 5(1):31.

9 Hills TT, Proto E, Sgroi D, Seresinhe CI (2019) Historical analysis of national subjective wellbeing using millions of digitized books. *Nature Human Behaviour* 3:1271–1275, URL http://dx.doi.org/10. 1038/s41562-019-0750-z.

10 Jacobs, A. M., Schuster, S., Xue, S. & Lüdtke, J. (2017). Whats in the brain that ink may character ....: A Quantitative Narrative Analysis of Shakespeares 154 Sonnets for Use in Neurocognitive Poetics, *Scientific Study of Literature 7:1, 4–51*. doi 10.1075/ssol.7.1.02jac.

11 Xue, S., Lüdtke, J., Sylvester, T., & Jacobs, A. M. (2019). Reading Shakespeare Sonnets : Combining Quantitative Narrative Analysis and Predictive Modeling — an Eye Tracking Study. *Journal of Eye Movement Research*, *12*(5). https://doi.org/10.16910/jemr.12.5.2

12 Hsu, C. T., Jacobs, A. M., Citron, F., & Conrad, M. (2015). The emotion potential of words and passages in reading Harry Potter - An fMRI study. *Brain and Language*, 142, 96-114.

13 Lüdtke, J., Meyer-Sickendiek, B., and Jacobs, A. M. (2014). Immersing in the stillness of an early morning: testing the mood empathy hypothesis in poems. *Psychol. Aesthet.* 8, 363–377. doi: 10.1037/a0036826

14 Xue S, Jacobs AM and Lüdtke J (2020) What Is the Difference? Rereading Shakespeares Sonnets — An Eye Tracking Study. *Front. Psychol.* 11:421. doi: 10.3389/fpsyg.2020.00421

15 Jacobs AM (2019) Sentiment Analysis for Words and Fiction Characters From the Perspective of Computational (Neuro-)Poetics. *Front. Robot. AI* 6:53. doi: 10.3389/frobt.2019.00053

16 Jakobson, R., and Lévi-Strauss, C. (1962). "Les chats" de charles baudelaire. *LHomme* 2, 5–21. doi: 10.3406/hom.1962.366446

17 Jockers, M. and Witten, D. (2010). A comparative study of machine learning methods for authorship attribution. *Literary and Linguistic Computing*, 25: 215–23

18 Kessler, B., Nunberg, G., & Schutze, H. (1997). Automatic detection of text genre. In *Proceedings of the 35th Annual Meeting of the Association for Computational Linguistics*. Madrid, Spain, pages 32–38.

19 Kim, E., Padó, S., & Klinger, R. (2017). Investigating the relationship between literary genres and emotional plot development. *Proceedings of the Joint SIGHUM Workshop on Computational Linguistics for Cultural Heritage, Social Sciences, Humanities and Literature*, 17–26.

20 Jacobs, A. M., & Kinder, A. (2019). Computing the Affective-Aesthetic Potential of Literary Texts, *Artifical Intelligence*, 1:1, 11–27; doi:10.3390/ai1010002

21 Bestgen, Y. (1994). Can emotional valence in stories be determined from words? *Cogn. Emot.* 8, 21–36. doi: 10.1080/02699939408408926

22 Westbury, C., Keith, J., Briesemeister, B. B., Hofmann, M. J. & Jacobs, A. M. (2015). Avoid violence, rioting, and outrage; approach celebration, delight, and strength: Using large text corpora to compute valence, arousal, and the basic emotions. *Quarterly Journal of Experimental Psychology* **68,** 1599–1622, doi: 10.1080/17470218.2014.970204.

23 Jacobs AM (2017) Quantifying the Beauty of Words: A Neurocognitive Poetics Perspective. *Front. Hum. Neurosci*. 11:622. doi: 10.3389/fnhum.2017.00622

24 Jacobs, A. M., & Kinder, A. (2018). What makes a metaphor literary? Answers from two computational studies, *Metaphor and Symbol*, 33:2, 85-100, DOI: 10.1080/10926488.2018.1434943

25 Klare, G. R. (1974–1975). Assessing readability. *Reading Research Quarterly*, **10**, 62-102.

26 Mikolov, T., Chen, K., Corrado, G., & Dean, J. (2013). Efficient estimation of word representations in vector space. Retrieved from https://arxiv.org/abs/1301.3781

27 Strobl, C., Malley, J., and Tutz, G. (2009). An introduction to recursive partitioning: rationale, application, and characteristics of classification and regression trees, bagging, and random forests. *Psychol. Methods* 14, 323–348. doi: 10.1037/a0016973

28 Mosteller, R. F. and Wallace, D. L. (1964). Inference and Disputed Authorship: The Federalist. Reading, MA: Addison-Wesley.

29 Eder, M. (2015). Does size matter? Authorship attribution, small samples, big problem. *Digital Scholarship in the Humanities*, 30(2): 167–82.

30 Jacobs AM (2015) Neurocognitive poetics: methods and models for investigating the neuronal and cognitive-affective bases of literature reception. *Front. Hum. Neurosci*. 9:186. doi: 10.3389/fnhum.2015.00186

31 Willems, R. & Jacobs, A.M. (2016). Caring about Dostoyevsky: The untapped potential of studying literature. *Trends in Cognitive Sciences*, 20, 243-245. https://doi.org/10.1016/j.tics.2015.12.009

32 Schroeder, S., Würzner, K. M., Heister, J., Geyken, A., & Kliegl, R. (2015). childLex: A lexical database of German read by children. *Behavior Research Methods*, *47*(4), 1085-1094.

33 Jacobs AM, Hofmann MJ and Kinder A (2016) On Elementary Affective Decisions: To Like Or Not to Like, That Is the Question. *Front. Psychol*. 7:1836. doi: 10.3389/fpsyg.2016.01836

34 Bird, S., Klein, E., & Loper, E. (2009). Natural language processing with Python. OReilly Media.

35 Schmid, H. (1994). Probabilistic Part-of-Speech Tagging Using Decision Trees. *Proceedings of International Conference on New Methods in Language Processing*, Manchester, UK.

36 Saltelli, A. (2002). Sensitivity analysis for importance assessment. *Risk Anal*. 22, 579–590. doi: 10.1111/0272-4332.00040

## Acknowledgements

None

## Author contributions

A.J. designed and performed the study. A.K. and A.J. performed the ML part and analysed the results. AJ drafted the manuscript, A.J. and A.K contributed both to the final version of the paper.

## Competing interests

The authors declare no competing interests.

## Additional information

**Supplementary information** for this paper will be made available at github or alternative outlets.

**Correspondence and requests for materials** should be addressed to A.J.